\documentclass[ conference]{./support/ieeeconf}
\usepackage{times}
\usepackage{amsmath}
\usepackage[pdftex]{graphicx}
\usepackage{caption}
\usepackage{subfiles} 
\usepackage{subfigure}
\usepackage{graphicx}
\usepackage{amsmath,amssymb,amsopn,amstext,amsfonts}
\usepackage{cancel}
\usepackage[space]{cite}
\usepackage{pdfsync}
\usepackage{balance}
\usepackage{color}
\usepackage{algorithm}
\usepackage{algorithmic}
\usepackage{url}
\usepackage{booktabs}
\usepackage{threeparttable}
\usepackage[linkcolor=black,citecolor=black,urlcolor=black,colorlinks=true]{hyperref}
\usepackage{multirow}
\usepackage[outdir=./epstopdf/]{epstopdf}
\usepackage{graphicx}
\usepackage{array}
\usepackage{verbatim}
\usepackage{makecell}
\usepackage{subcaption}
\usepackage{indentfirst} 
\IEEEoverridecommandlockouts

\graphicspath{{./figure/}}

\DeclareGraphicsExtensions{.pdf,.png,.jpg,.eps,.PNG}
\title{\LARGE \bf RL-OGM-Parking: Lidar OGM-Based Hybrid Reinforcement Learning \\ Planner for Autonomous Parking }
\author{Zhitao Wang$^{\dag}$, Zhe Chen$^{\dag}$, Mingyang Jiang, Tong Qin, and Ming Yang
        \thanks{All authors are with the Global Institute of Future Technology, Shanghai Jiao Tong University, Shanghai, China.
		{\tt\small  \{wzt122196, Zhe{\_}Chen,jiamiya, qintong, mingyang\}@sjtu.edu.cn}.   
	}
 \thanks{$\dag$ means contributing equally.}
 }

\begin{document}

\maketitle
\thispagestyle{empty}
\pagestyle{empty}

\begin{abstract}
Autonomous parking has become a critical application in automatic driving research and development. 
Parking operations often suffer from limited space and complex environments, requiring accurate perception and precise maneuvering. 
Traditional rule-based parking algorithms struggle to adapt to diverse and unpredictable conditions, while learning-based algorithms lack consistent and stable performance in various scenarios. 
Therefore, a hybrid approach is necessary that combines the stability of rule-based methods and the generalizability of learning-based methods.
Recently, reinforcement learning (RL) based policy has shown robust capability in planning tasks.
However, the simulation-to-reality (sim-to-real) transfer gap seriously blocks the real-world deployment. 
To address these problems, we employ a hybrid policy, consisting of a rule-based Reeds-Shepp (RS) planner and a learning-based reinforcement learning (RL) planner.
A real-time LiDAR-based Occupancy Grid Map (OGM) representation is adopted to bridge the sim-to-real gap, leading the hybrid policy can be applied to real-world systems seamlessly.  
We conducted extensive experiments both in the simulation environment and real-world scenarios, and the result demonstrates that the proposed method outperforms pure rule-based and learning-based methods.
The real-world experiment further validates the feasibility and efficiency of the proposed method.
\end{abstract}

\section{Introduction}
With the rapid advancement of intelligent driving technology, autonomous parking has emerged as a critical area of research within automatic driving. 
Parking tasks require precise vehicle maneuvers in narrow, congested, and dynamic environments, presenting significant challenges for perception, planning, and control systems\cite{li2024parkinge2e}. 
Traditional parking algorithms, based on rule-based control models, struggle to adapt to complex and unpredictable real-world scenarios, limiting their generalizability. 
Besides, reinforcement learning (RL) has gained traction in autonomous parking due to its ability to optimize strategies through interaction. 
In other words, integrating the reliability found in rule-based techniques with the adaptability of learning-driven strategies is essential for developing a more robust methodology.
However, RL-based models trained in simulations face challenges in real-world deployment due to the gap between the observations in simulation environment and reality.
Specifically, current simulation environments rely on manually defined rules for observation rendering, which fail to accurately replicate the abundant visual information and unpredictable noises of the real world. This makes deploying models trained in such simulations directly into the real world suffer from \cite{robotics_sim2real_yu2024}, which remains a critical issue for practical application.

\begin{figure}[t]
	\centering
        \includegraphics[width=\linewidth]{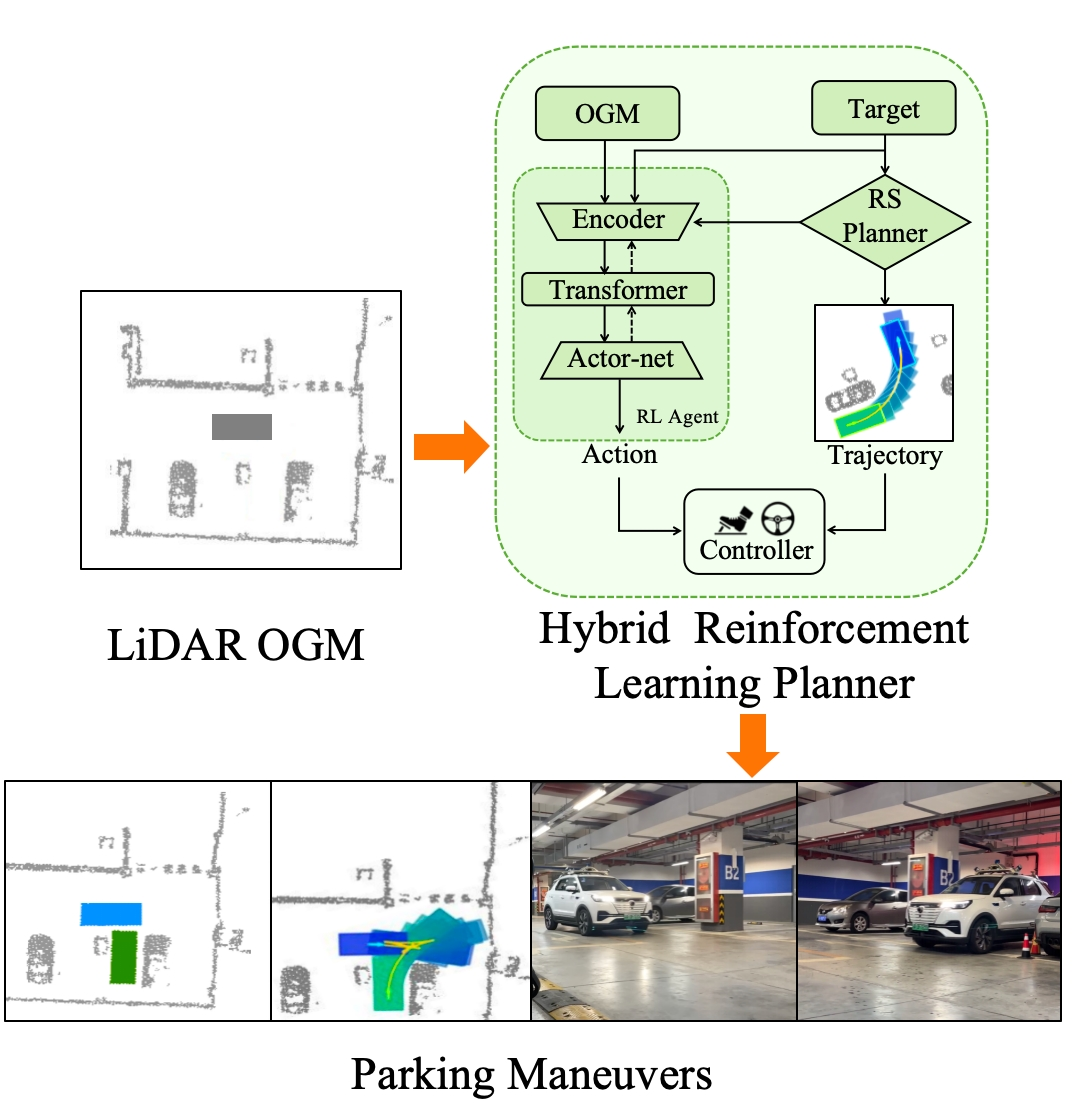}
        \vspace{-0.6cm}
	\caption{The figure shows the brief structure of the overall workflow. The environmental information is transformed into Occupancy Grid Map (OGM) and sent into the Hybrid Reinforcement Learning Planner for parking maneuvers.}
	\label{fig:abs}
       \vspace{-0.6cm}
\end {figure}

{T}{o} address this challenge, this paper proposes an innovative RL-based autonomous parking approach, achieving a fully closed-loop system from perception to planning and control. 
We adopt Occupancy Grid Map (OGM) representation through LiDAR-based real-time perception, converting complex environmental information into occupancy grid maps for both simulation environment construction and real-time inference. 
Depth-based OGM provides perception results with a consistent observation input of RL agent, which effectively reduces the gap between simulation environments and real-world scenarios. 

{F}{urthermore}, this paper adopts a hybrid reinforcement learning planner\cite{jiang2024hope}, combining a data-driven RL planner with the rule-based Reeds Shepp (RS)\cite{RS_curve1990} planner to achieve complementary advantages. 
The RL model dynamically adjusts the vehicle's pose based on perception results. 
Meanwhile, the RS planner generates the trajectories leading to the target position, which guarantees high decision-making speed and stability, significantly enhancing the efficiency of training and parking maneuvers.
We deploy the proposed system on a real-vehicle experiment platform and validate its capability through both simulation and real-vehicle experiments in different scenarios. 
The results demonstrate that the proposed parking system can effectively perform parking maneuvers in real-world complex environments while maintaining strong adaptability and generalization capabilities.

The contributions of this paper are summarized as follows:
\begin{itemize}
	\item We propose a hybrid reinforcement learning planner that takes Lidar
 Occupancy Grid Maps (OGM) as the input in both training and inference stages. The OGM presentation effectively reduces the sim-to-real gap by aligning the simulation and real-world perception input.
	\item We deployed the hybrid reinforcement learning planner on \textbf{real vehicles} for testing and verified the parking capability, feasibility, and generalizability across various \textbf{real-world scenarios}, providing a valid solution for the application of reinforcement learning in autonomous parking.
\end{itemize}   

\section{literature review}

\subsection{Reinforcement Learning for Autonomous  Parking}
In the field of automatic parking, traditional parking algorithms often rely on predefined rules or manually designed controllers, making it difficult to handle complex and dynamic real-world scenarios. 
To address these challenges, reinforcement learning (RL) has gradually emerged as an innovative approach in autonomous parking. 
Through continuous interaction and self-learning, RL algorithms can optimize decision-making strategies in dynamic environments, enabling more intelligent and flexible parking maneuvers.

In the early stages of artificial intelligence development, Maravall et al.\cite{maravall2003automatic} proposed a theoretical framework for autonomous parking using artificial neural networks (ANN), designing both model-free and model-based RL parking networks. 
In recent years, many studies have also explored the use of RL algorithms for trajectory planning in parking scenarios. 
Thunyapoo et al.\cite{thunyapoo2020self} proposed an autonomous parking approach using Proximal Policy Optimization (PPO) in a simulation environment. 
Takehara et al.\cite{takehara2021autonomous} combined sensor data and introduced a purely visual deep reinforcement learning (DRL) parking solution based on the Unity simulator. 
Zhang et al.\cite{zhang2020data} extended RL to the data generation phase by designing reward functions to filter the generated data, optimizing model performance, and reducing dependence on human data. Additionally, Zhang et al.\cite{zhang2019e2e} integrated real sensor data and proposed an end-to-end RL algorithm that uses parking space tracking for perception, which was deployed on an experimental vehicle. This approach improved performance in real-world deployments through a multi-stage training strategy.

Some studies have also combined rule-based non-learning trajectory planners with learning-based planners, proposing hybrid RL parking algorithms. 
For instance, Shi et al.\cite{shi2023model} introduced a hybrid parking algorithm that uses a rule-based Model Predictive Control (MPC) planner for parking space searching and a RL algorithm for parking maneuvers. 
Jiang et al. \cite{jiang2024hope} proposed Hybrid Policy Path Planner (HOPE), which integrates RL algorithms with RS trajectory planners during the parking phase, enhancing parking capability and training efficiency.

\subsection{Simulation-to-Reality Transfer in Reinforcement Learning}

Sim-to-Real (Simulation-to-Reality) refers to the process of applying reinforcement learning strategies or models, trained in simulation environments, to real-world scenarios\cite{zhao2020sim}. 
In reinforcement learning, simulation environments offer abundant, low-cost, and limitless training data, facilitating rapid iterations and strategy optimization. 
However, due to discrepancies of sensor inputs, physics simulation, and sensor and control delays\cite{robotics_sim2real_yu2024} between simulation and reality, strategies trained in simulations often experience performance degradation when deployed in the real world, a phenomenon known as the “Sim-to-Real Gap".

To address this issue, Sim-to-Real methods aim to bridge the gap between simulation and reality through various techniques, ensuring robust performance in real-world applications. 
Some approaches reduce the sim-to-real gap by minimizing the differences between the simulator and reality. 
Tobin et al.\cite{tobin2017domain} proposed Domain Randomization, which enhances robustness in real environments by randomizing simulator parameters such as colors, textures, and dynamics. 
Bousmalis et al.\cite{bousmalis2018using} introduced Domain Adaptation, aligning the feature spaces of the source and target domains to reduce differences and improve the model's adaptability in real environments.


Inspired by the hybrid policy proposed by Jiang et al.\cite{jiang2024hope}, we propose an innovative hybrid reinforcement learning parking algorithm that uses LiDAR Occupancy Grid Maps (OGM) as input. This algorithm improves upon the original hybrid strategy combining the Reeds-Shepp (RS) planner and reinforcement learning planner by leveraging LiDAR OGM to unify complex environmental information into a consistent input format, effectively reducing the Sim-to-Real Gap between simulation and reality.

\begin{figure*}[t]
	\centering
	\includegraphics[width=\textwidth]{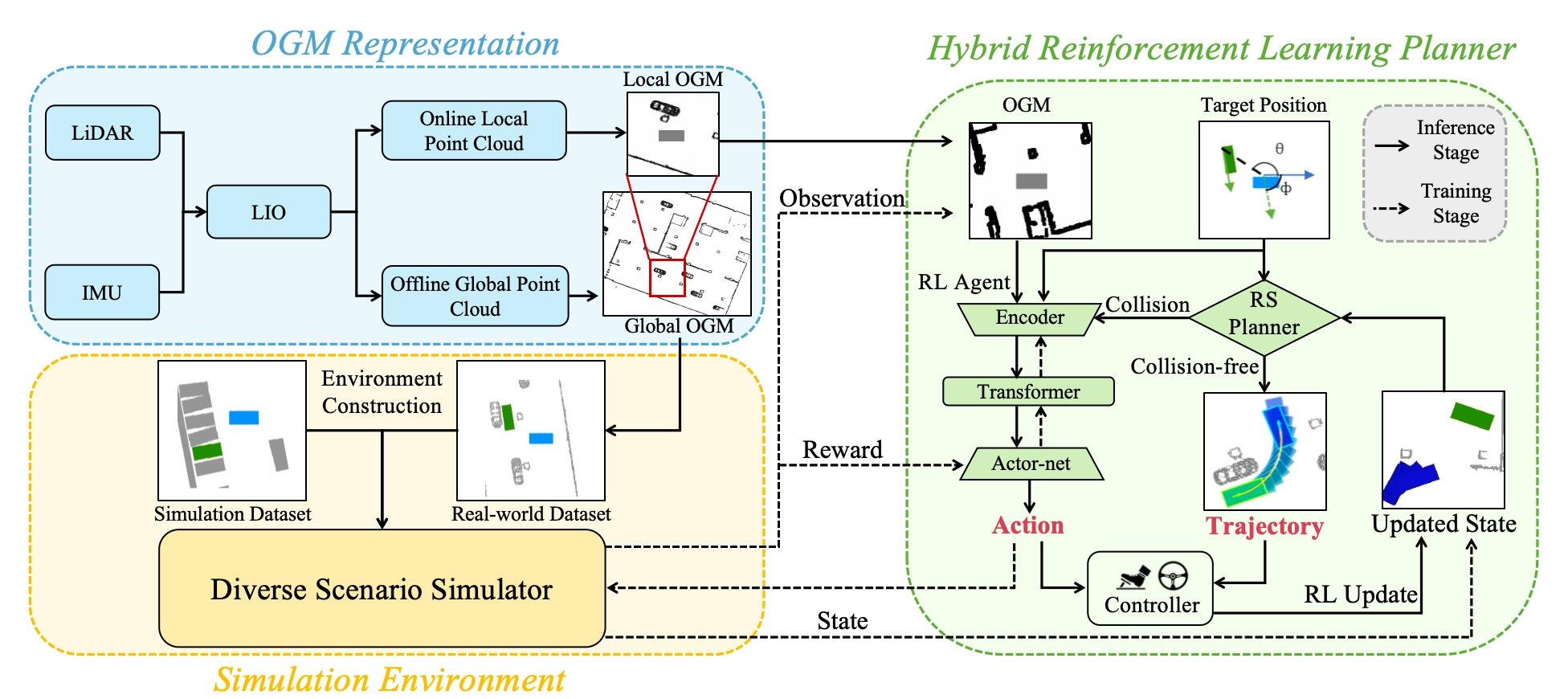}
	\caption{
		The figure shows the overall workflow. The proposed integrated hybrid reinforcement learning system for autonomous parking leverages OGM representations derived from LiDAR and IMU data. 
  The global OGM aids in creating diverse simulation scenarios, while the local OGM is used for inference in the hybrid reinforcement learning network. 
  The hybrid reinforcement learning planner combines a rule-based Reeds-Shepp (RS) planner with a learning-based RL planner, taking OGMs and target positions as inputs to generate actions and trajectories for precise parking maneuvers.
		}
	\label{fig:framework}
       \vspace{-0.5cm}
\end{figure*}

\section{methodology}
\subsection{Overview of Architecture}
The overall framework for path planning based on the hybrid reinforcement learning planner is shown in Fig. \ref{fig:framework}. 
To maintain consistent model performance across virtual and real-world environments and eliminate the sim-to-real gap, we adopt the OGM representation as the input. 
The LiDAR data is converted into Occupancy Grid Maps (OGM).
The global OGM is used to generate simulation scenarios during the training stage, while the local OGM is used as the real-time perception for inference.
The hybrid reinforcement learning planner combines a rule-based Reeds-Shepp (RS) planner with a learning-based planner, taking OGMs and target positions as inputs.
The RL model generates actions to adjust vehicle pose while the RS planner generates feasible trajectories to finish the parking process.

\subsection{OGM Representation}
For perception, we propose the Occupancy Grid Map (OGM) representation to express the environment, capturing obstacle information and free space. 
This module takes multiple frames of LiDAR point clouds and IMU data as input. 
During OGM generation, point clouds from ceilings and ground surfaces significantly interfere with perception results. 
Therefore, a filter is applied to the input point clouds to remove these points based on their relative height to the LiDAR sensor. 
After filtering, the LiDAR point cloud is processed by a LiDAR-IMU odometry (LIO) system\cite{bai2022faster}, where point-plane features are extracted from the point cloud and combined with IMU measurements for pose estimation, yielding the final pose estimation.

Once accurate pose estimation, corresponding ego-to-global transformation matrix $\mathcal{T}_t$, and filtered point clouds $\mathcal{C}_t$ are obtained, two processing methods are applied depending on the use case.
In the training phase for environment generation, the point cloud is registered into a global map based on the pose estimation, creating a global LiDAR point cloud (PCD) map $\mathcal{M}_G$ by
\begin{equation}
    \mathcal{M}_G=\sum_{t=1}^{t} \mathcal{T}_t \mathcal{C}_t.
\end{equation}
This map is then processed by further filters and splatted into a 2D plane, generating a global occupancy grid map.

During the inference phase, multiple frames of point clouds and pose estimation results are fused into a keyframe to generate a local LiDAR point cloud map $\mathcal{M}_L$ by 
\begin{equation}
\mathcal{M}_L = \sum_{t}^{t_k} \mathcal{T}_{t_k}^{-1} \mathcal{T}_t \mathcal{C}_t,
\end{equation}
where $t_k$ is the time step of the keyframe.
 
The local LiDAR point cloud map is also processed by further filters and generates a real-time local occupancy grid map for inference.
The OGM representation ensures input consistency across virtual training and real-time perception, reducing the sim-to-real gap commonly encountered in RL model deployment and improving model generalization and practical application.

\subsection{Simulation Environment}
The simulator leverages the Gym toolkit to build the reinforcement learning environment and employs Pygame for visual rendering, simulating a bird's-eye view (BEV) parking scenario. 
This environment provides a standardized interface for reinforcement learning, facilitating the interaction between the parking agent and the simulation environment.
We utilize two datasets: \textbf{Simulation dataset} and \textbf{Real-World dataset} to construct the simulation environment, ensuring the model can be trained in various scenarios to maximize the parking capability and generalizability.
\begin{itemize}
    \item \textbf{Simulation dataset:} These maps are generated in simulation environment using predefined rules and randomized algorithms to create starting positions, target locations, and obstacles, incorporating multiple scenarios including 20 parallel parking scenarios and 50 perpendicular parking scenarios.
    \item \textbf{Real-World dataset:} These maps are generated by the global OGM from the OGM representation module, which collects LiDAR data in several real-world underground garages.
\end{itemize}

\subsection{Hybrid Reinforcement Learning Planner}
To enable efficient path planning in complex parking scenarios, we employed a hybrid approach that integrates reinforcement learning with a traditional rule-based Reeds-Shepp (RS) planner, utilizing the complementary strengths of both methods for improving planning performance.

\subsubsection{Reeds-Shepp Path Planner}
The RS curve\cite{RS_curve1990} provides reliable preliminary path planning. 
RS curve is designed to generate the shortest path between two positions, taking into account the vehicle’s minimum turning radius and straight-line segments. 
It has been mathematically proven that the shortest path belongs to one of 48 types of curves, which can be represented by the 9 expressions.

\subsubsection{Reinforcement Learning Path Planner}
Reinforcement learning (RL) continually interacts with a simulation environment, enabling effective exploration and complex decision-making through self-updating. 
In our framework, the RL problem is modeled as a Markov Decision Process (MDP) defined by a state space $S$, action space $A = \{(v, \delta)\}$, state transition probability $p$, and reward function $r$. 
The action space consists of vehicle control inputs: velocity $v$ and steering angle $\delta$.

We utilize the Soft Actor-Critic (SAC)\cite{SAC2018} algorithm, which is a maximum entropy reinforcement learning method that maximizes the cumulated reward as well as policy entropy. The loss function of SAC can be expressed as:

\begin{equation}
    L(\theta) = \mathbb{E}_{s_t \sim D} [\alpha H(\pi(\cdot | s_t)) - Q(s_t, a_t)]
\end{equation}
where $\theta$ refers to the parameters of the policy $\pi$, and $Q(s_t, a_t)$ is the action-value function that estimates the expected cumulative reward of taking action $a_t$ in state $s_t$. The entropy term $H(\pi(\cdot | s_t)) = \log \pi_{\theta}(a_t | s_t)$ incentivizes policy diversity and exploration, allowing the agent to learn and adapt to various strategies effectively.

To prevent unexpected collisions and improve training efficiency, we integrated action mask into the reinforcement learning path planner. 
The action mask restricts the range of feasible actions by determining the maximum safe velocity for each steering angle, based on the vehicle's current state and environmental constraints.

\subsubsection{Hybrid Reinforcement Learning Planner}

We employ a hybrid reinforcement learning planner in our autonomous parking system, combining rule-based and learning-based path planning. 
The planner processes OGMs from real perception or simulation environment, along with the target position relative to the ego vehicle. 

To improve the efficiency of exploration, the agent utilizes the RS planner at each step to check whether a collision-free and feasible RS curve exists. 
If no such curve exists, which is often the case in the initial stages of the parking process, the agent relies on the RL planner's ability to interact with the environment for exploration. The RL planner takes OGMs and the target as inputs and generates candidate actions, which are constrained to collision-free regions by an action mask, ensuring safer decisions. 

During training, these actions are passed into the simulation environment, returning updated states, observations, and rewards. The new state is then provided to the agent for the next iteration. 
In real-world deployment, actions are sent directly to the chassis controller to adjust the vehicle’s position.

This hybrid planner utilizes the exploration and adaptability of the RL planner while benefiting from the stability and efficiency of the RS planner.
This hybrid policy ensures the agent learns to adjust the vehicle's pose, increasing the chances of successful parking trajectories.

\section{Experiments}
To evaluate the performance of the proposed method, we conducted experiments in both simulation and real-world environments.
\subsection{Simulation Evaluation}

\subsubsection{Implementation Details}
We designed various parking scenarios according to the difficulty levels, including Sim-Normal, Sim-Complex, and Real-World.
The simulation environment generates occupancy grid maps (OGMs), which serve as inputs to the hybrid reinforcement learning algorithm. 

\begin{itemize}
    \item \textbf{Sim-Normal:} Scenarios are generated using Simulation dataset with a limited number of obstacles.
    \item \textbf{Sim-Complex:} Scenarios are derived from Simulation dataset featuring an increased number of obstacles and narrow space.
    \item \textbf{Real-World:} Scenarios are based on Real-World datasets, characterized by a high density of obstacles and restricted free space.
\end{itemize}

In each scenario, the system’s planning accuracy and robustness were tested and compared with the following baseline methods:
\begin{itemize}
    \item \textbf{Hybrid A*}: Hybrid A*\cite{hybridAstar} is a path planning algorithm that merges A* efficiency with smooth motion and steering constraints. To make it more efficient, the latest version is embedded with RS curves, which is this paper utilized.
    \item \textbf{SAC (Soft Actor-Critic)}: SAC \cite{SAC2018} is a reinforcement learning algorithm that maximizes both reward and action diversity by optimizing a balance between expected return and entropy.
    \item \textbf{PPO (Proximal Policy Optimization)}: PPO\cite{PPO2017} is a reinforcement learning algorithm that improves training stability by limiting large policy updates, ensuring more reliable learning.
\end{itemize}

We also compared the parking capability of rule-based Hybrid A* and hybrid reinforcement learning methods in three different real-world garage scenarios of different difficulty levels. 

\subsubsection{Simulation Experiment Results}
    We conducted tests in several typical parking scenarios of different difficulty levels, including normal, complex, and real-world scenarios. The test metrics are as follows:
\begin{itemize}
    \item \textbf{Parking Success Rate (PSR)}: The ratio of successful parking tasks to the total number of tasks attempted. 
    A parking attempt is deemed successful if the vehicle reaches the target position without collision and within the parking spot.
    \item \textbf{Number of Gear Shifts (ANGS)}: The average number of gear shifts made by the vehicle during the parking process. 
    This metric indicates the smoothness and efficiency of parking maneuvers, with fewer shifts generally being preferable.
    \item \textbf{Path Length (PL)}: The average length of all successful parking attempts. 
    The lower values indicate more efficient parking capabilities.
\end{itemize}

\begin{table}[t]
    \centering
    \caption{Comparison in Different Simulation Scenarios.}
    \setlength{\tabcolsep}{4pt} 
    \small
    \resizebox{0.5\textwidth}{!}{ 
        \begin{tabular}{l|c|ccc} 
            \toprule
            \textbf{Scenario} & \textbf{Method} & \textbf{PSR (\%) $\uparrow$} & \textbf{ANGS (times) $\downarrow$} & \textbf{PL (m) $\downarrow$} \\
            \midrule
            \multirow{4}{*}{Sim-Normal}    
            & Hybrid A*      & 79.27  & 1.7 & \textbf{18.7} \\
            & SAC            & 53.46  & 4.1 & 22.7 \\
            & PPO            & 72.17  & 5.7 & 24.2 \\
            & Hybrid RL (ours)       & \textbf{99.33} & \textbf{1.5}  & 20.3 \\
            \midrule
            \multirow{4}{*}{Sim-Complex} 
            & Hybrid A*      & 81.8   & 2.4 & 27.7 \\
            & SAC            & 37.5   & 5.5 & 32.9 \\
            & PPO            & 38.2   & 7.8 & 37.4 \\
            & Hybrid RL (ours)      & \textbf{97.7}  & \textbf{1.9}  &  \textbf{23.6} \\
            \midrule
            \multirow{4}{*}{Real-World} 
            & Hybrid A*      & 77.6   & 4.2 & 42.5 \\
            & SAC            & 23.1   & 7.4 & 54.2 \\
            & PPO            & 21.8   & 10.1 & 51.7 \\
            & Hybrid RL (ours)      & \textbf{87.2}  & \textbf{2.6} & \textbf{28.3} \\
            \bottomrule
        \end{tabular}
    }
    \label{compare_table_sim1}
\vspace{-1cm}
\end{table}

According to the experimental results presented in Table \ref{compare_table_sim1}, the hybrid reinforcement learning planner consistently outperforms others, particularly in parking success rate and maneuver smoothness.


In the \textbf{normal scenario}, our approach led with the highest parking success rate(99.33\%) and the lowest average number of gear shifting(1.5), proving its superior navigation and parking capabilities. Despite Hybrid A* showing slightly better path efficiency in path length (18.7 m), our approach offered the most balanced performance overall.
In the \textbf{complex scenario}, our approach maintained dominance with a parking success rate of 97.7\%, far outperforming other methods. It exhibited robust control with the lowest average number of gear shifting(1.9) even in challenging conditions. Additionally, our approach achieved the shortest path length of 23.6 m, demonstrating not only efficiency but also precision in navigation.
In the \textbf{real-world scenario}, it continued to excel, achieving the best parking success rate(87.2\%), the shortest path length (28.3 m), and minimal average number of gear shifting(2.6), demonstrating high efficiency and control under practical conditions. This robust performance across various settings underscores our approach's potential in autonomous parking technologies.

\begin{figure}[t]
	\centering
        \includegraphics[width=\linewidth]{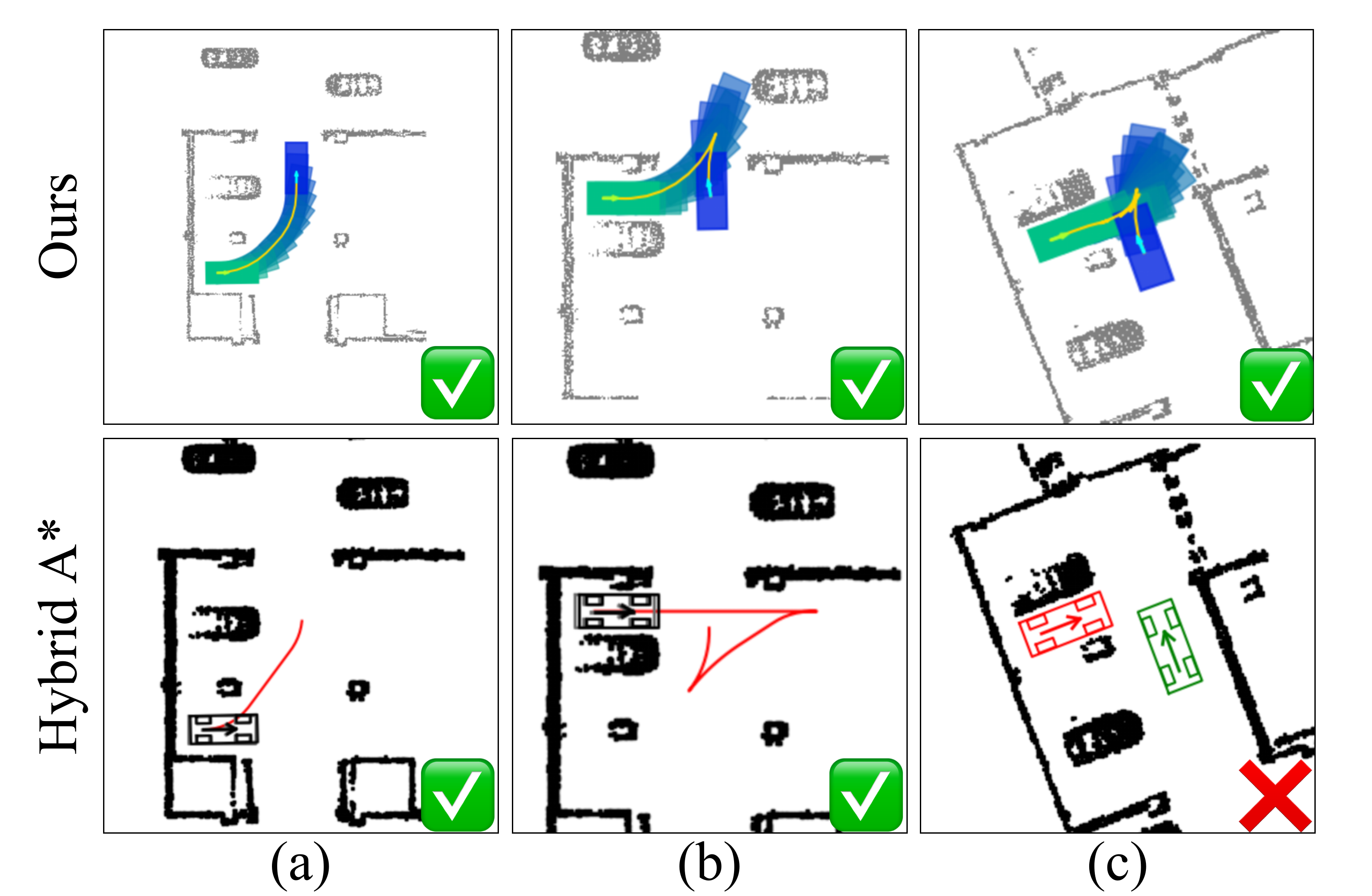}
	\caption{The path comparison of Hybrid A* and our proposed method in three different scenarios: (a) Simple, (b) Normal, (c) Extreme.}
	\label{compare_fig_sim2}
        \vspace{-0.3cm}
\end {figure}

\begin{table}[t]
\vspace{0.4cm}
    \centering
    \caption{Comparison in Real-World Dataset}
    \resizebox{0.4\textwidth}{!}{
        \begin{tabular}{c|c|c|c}
        \toprule
        \textbf{Scenario} & \textbf{Method} & \textbf{NGS(times)$\downarrow$} & \textbf{PL (m)$\downarrow$} \\
        \midrule
        \multirow{2}{*}{Simple} & Hybrid A* & 0 & 11.6 \\
                                & Hybrid RL (ours) & 0 & 12.63 \\
        \midrule
        \multirow{2}{*}{Normal} & Hybrid A* & 2 & 26.40 \\
                                & Hybrid RL (ours) & 1 & 14.39 \\
        \midrule
        \multirow{2}{*}{Extreme} & Hybrid A* & fail & fail \\
                                 & Hybrid RL (ours) & 6 & 16.31 \\
        \bottomrule
        \end{tabular}
    }
    \label{compare_table_sim2}
\vspace{-1cm}
\end{table}

As for the comparison in different real-world garage scenarios, the results are shown in Table. 
We classify the scenarios into simple, normal, and extreme according to the operation difficulty and take \textbf{Number of Gear Shifting (NGS)} and \textbf{Path Length (PL)} as metrics. \ref{compare_table_sim2} and the path comparison results are shown in Fig. \ref{compare_fig_sim2}. 
In the \textbf{simple scenario}, both methods exhibit stable performance, with Hybrid A* demonstrating more efficient parking maneuvers. 
In the \textbf{normal scenario}, our proposed method consistently maintains stable and efficient parking capabilities, whereas Hybrid A* generates a more complex path with more gear shifts. 
In the \textbf{extreme scenario}, our method performs precise adjustments in narrow spaces with a minimal number of gear shifts. 
In contrast, Hybrid A* fails to find a valid path due to the excessively complex dynamic environment and its reliance on open spaces for effective planning.

\begin{figure*}[t]
	\centering
	\includegraphics[width=0.95\textwidth]{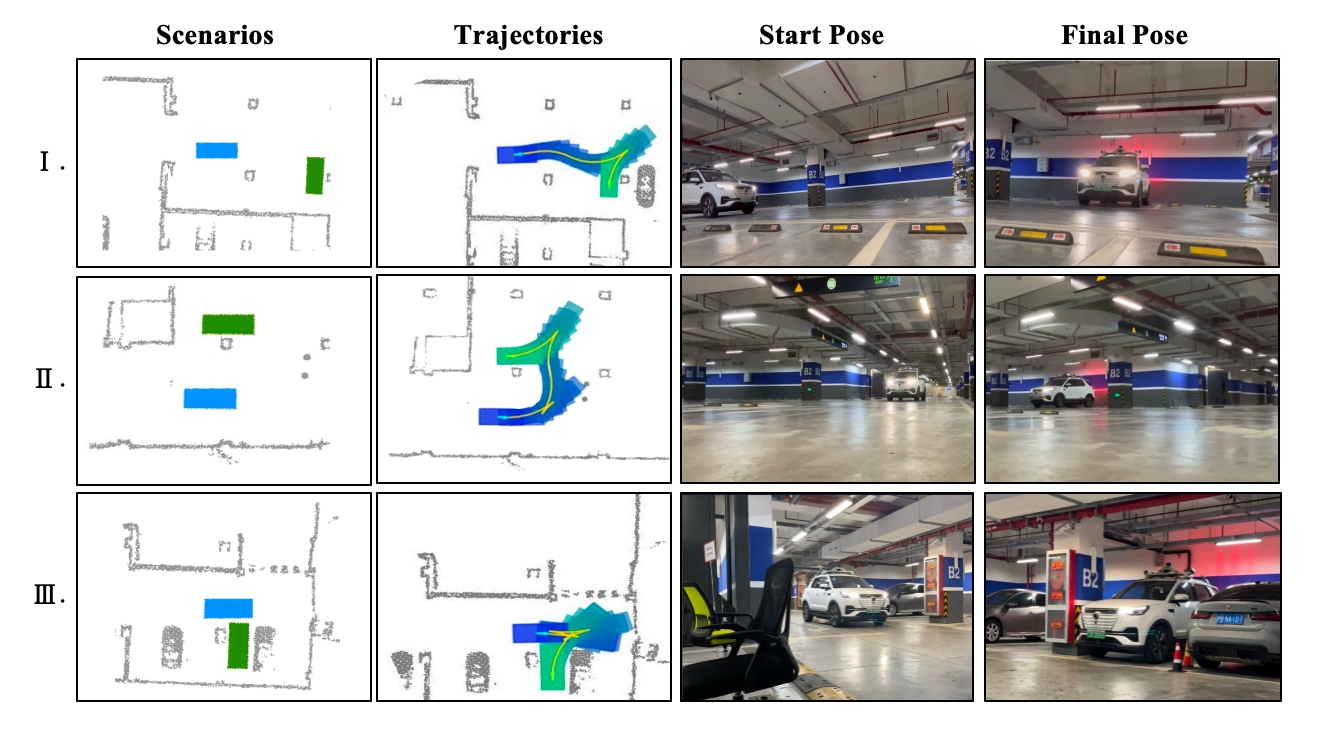}
        \vspace{-0.5cm}
	\caption{
		Figures shows the real-world experiment scenarios and results. The experiments were conducted in three complex scenarios: \uppercase\expandafter{\romannumeral1}. Long-Distance perpendicular parking, \uppercase\expandafter{\romannumeral2}. Long-Distance parallel parking, \uppercase\expandafter{\romannumeral3}. Narrow Dead-End parking. Photos illustrate the start and final pose of the vehicle.
  Details can be found in the supplementary video.
	}
	\label{real_result_fig}
 \vspace{-0.5cm}
\end{figure*}

\subsection{Real-World Evaluation}
\subsubsection{Implementation Details}
In real-world deployment, the hybrid reinforcement learning planner was integrated into a commercial vehicle experimental platform, equipped with four LiDAR sensors and an Inertial Measurement Unit (IMU). 
The hybrid model was trained with 2 mentioned datasets for 8 hours on workstation with NVIDIA RTX 4090 GPU.
The algorithm was implemented on an Intel NUC system, featuring an NVIDIA RTX 2060 GPU and an Intel Core i7 processor. 
During practical inference, the algorithm achieved an average inference time of $17.2 ms$, demonstrating its capability to support real-time decision-making and vehicle control.
The vehicle generated OGM representations from real-time perception data and utilized the pre-trained model for path planning. 
Tests were conducted in three challenging scenarios within an underground parking garage: 
\begin{itemize}
    \item \textbf{Long-Distance Perpendicular Parking (Scenario \uppercase\expandafter{\romannumeral1}): } The vehicle approaches the parking space from a significant distance while oriented perpendicularly to the slot.
    \item \textbf{Long-Distance Parallel Parking (Scenario \uppercase\expandafter{\romannumeral2}):} The vehicle is positioned on the road far from the parking space and the initial pose is aligned parallel to the slot.
    \item \textbf{Narrow Dead-End Parking (Scenario \uppercase\expandafter{\romannumeral3}):} The parking space is located at the end of a narrow road with obstacles on both sides of the parking slot.
\end{itemize}

\subsubsection{Real-World Experiment Results}
In real-world experiments, we conduct 20 parking attempts in each scenario. 
To quantitatively measure the performance of the hybrid reinforcement learning planner, we utilize test metrics as follows: 
\begin{itemize}
   
    \item \textbf{Average Operation Time (AOT) }: The average time cost of parking operations.
    
    \item \textbf{Average Number of Gear Shifting (ANGS)}: The average number of gear shifting operations.
\end{itemize}

\begin{table}[t]
    \centering
    \caption{Quantitative Results in Real-World Scenarios}
    \begin{tabular}{c|c|c|c}

    \toprule
    \textbf{Scenario} & \textbf{PSR(\%) $\uparrow$} & \textbf{AOT(s) $\downarrow$} & \textbf{ANGS(times) $\downarrow$} \\
    \midrule
    Long-Dis. Perpendicular &  $100$ & 35 & 1 \\
    Long-Dis. Parallel      &  $85$ & 59 & 1 \\
    Narrow Dead-End         &  $60$  & 85 & 8.5 \\
    \midrule
    \end{tabular}
    \vspace{-1cm}
\end{table}

The results and scenarios of real-world experiments are shown in Fig. \ref{real_result_fig}. In real-world experiments, the hybrid model exhibited exceptional performance across all three complex parking scenarios. 
In both the Long-Distance Perpendicular and Long-Distance Parallel parking scenarios, the model effectively utilized real-time perception to optimize the use of available free space, achieving a high success rate in parking operations. 
Notably, all successful trials were completed within one minute, with only a single gear-shifting maneuver, demonstrating the model's precise control and parking efficiency. 
In the Narrow Dead-End scenario, the model leveraged accurate LiDAR-based perception to maximize the use of limited free space, enabling highly precise adjustments to the vehicle's position. 
These results highlight the model's robust ability to handle constrained environments while maintaining high levels of accuracy and efficiency.

\section{Conclusion}
In this paper, we proposed a hybrid reinforcement learning planner for autonomous parking, utilizing real-time Occupancy Grid Maps (OGM) for perception. 
The hybrid model combines the rule-based RS planner and a learning-based RL planner, taking OGMs as input both for training and inference. 
This method effectively addresses the sim-to-real transfer gap, enabling precise and efficient parking maneuvers in real-world scenarios. 
We evaluate the method across a range of simulation and real-world parking scenarios, demonstrating its excellent capability in handling complex parking environments. 
Our future work will focus on end-to-end models and improving simulators to enhance the generalization and parking performance. 
We believe that our work would offer insights and practice experience to advance the field of automatic driving.

\clearpage

\bibliography{reference.bib}

\end{document}